\documentclass{article}

\usepackage{PRIMEarxiv}

\usepackage[utf8]{inputenc}
\usepackage[T1]{fontenc}

\usepackage{hyperref}
\usepackage{url}
\usepackage{booktabs}
\usepackage{amsfonts}
\usepackage{nicefrac}
\usepackage{microtype}
\usepackage{float}

\usepackage{amsmath}
\usepackage{subcaption}
\usepackage{multirow}
\usepackage{graphicx}
\usepackage[square,numbers,sort&compress]{natbib}
\usepackage{placeins}
\usepackage{enumitem}
\usepackage{makecell}

\usepackage{caption}

\usepackage[normalem]{ulem}

\usepackage{xcolor}

\graphicspath{{./}{media/}}

\usepackage{fancyhdr}
\newcounter{corrfn}
\newcommand{\corrthanks}[1]{%
  \thanks{#1}\setcounter{corrfn}{\value{footnote}}%
}
\newcommand{\corrmark}{\footnotemark[\value{corrfn}]}

\title{Zero-Shot Stance Detection in the Wild: Dynamic Target Generation and Multi-Target Adaptation}

\author{
Aohua Li\textsuperscript{1,2},
Yuanshuo Zhang\textsuperscript{1,2},
Ge Gao\textsuperscript{2,4}\corrthanks{Corresponding authors: Ge Gao (Email: nmwgg@163.com), Bo Chen (Email: chenbomuc@muc.edu.cn).},
Bo Chen\textsuperscript{1,2,3}\corrmark,
Xiaobing Zhao\textsuperscript{1,2}
\\
\textsuperscript{1}School of Information Engineering, Minzu University of China, Beijing 100081, China\\
\textsuperscript{2}National Language Resource Monitoring and Research Center of Minority Languages, Beijing 100081, China\\
\textsuperscript{3}Institute of National Security, Minzu University of China\\
\textsuperscript{4}School of Minority Languages and Literatures, Minzu University of China, Beijing 100081, China\\
}

\begin{document}
\maketitle

\begin{abstract}
Current stance detection research typically relies on predicting stance based on given targets and text. However, in real-world social media scenarios, targets are neither predefined nor static but rather complex and dynamic. To address this challenge, we propose a novel task: zero-shot stance detection in the wild with Dynamic Target Generation and Multi-Target Adaptation (DGTA), which aims to automatically identify multiple target-stance pairs from text without prior target knowledge. We construct a Chinese social media stance detection dataset and design multi-dimensional evaluation metrics. We explore both integrated and two-stage fine-tuning strategies for large language models (LLMs) and evaluate various baseline models. Experimental results demonstrate that fine-tuned LLMs achieve superior performance on this task: the two-stage fine-tuned Qwen2.5-7B attains the highest comprehensive target recognition score of 66.99\%, while the integrated fine-tuned DeepSeek-R1-Distill-Qwen-7B achieves a stance detection F1 score of 79.26\%. The dataset and models are publicly available\footnote{\url{https://github.com/17Huaaa/DGTA-stance-detection/}}.
\end{abstract}


\section{Introduction}
Stance detection aims to identify an author’s attitudinal tendency toward a specific target, typically categorized as support, against, or neutral \cite{aldayel2021stance,2016SemEval,li-caragea-2019-multi}. Most existing work has focused on predefined targets and achieved notable progress in this setting \cite{kuccuk2020stance,siddiqua2019tweet}.

\begin{figure}[htbp]
  \centering
  \includegraphics[width=1.0\linewidth,height=0.15\textheight,keepaspectratio]{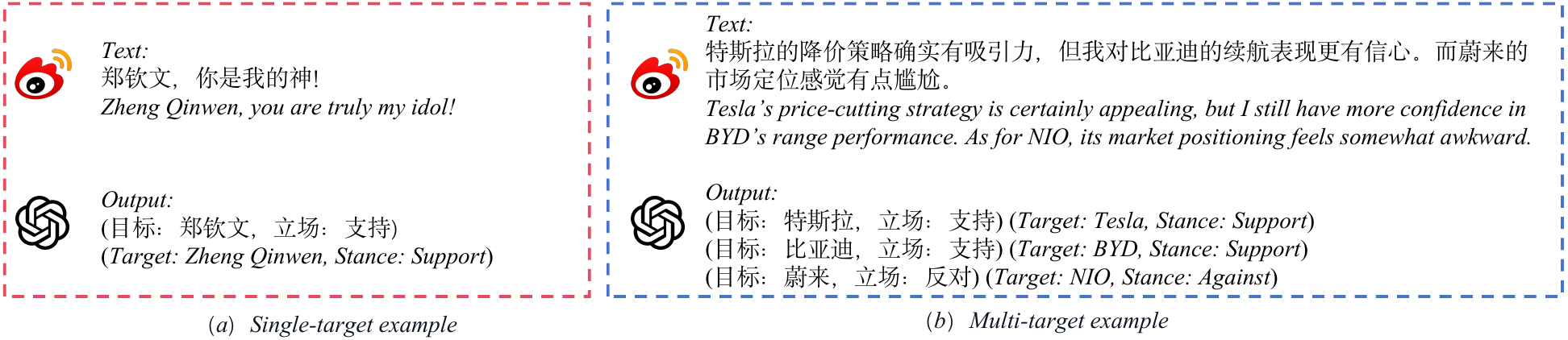}
  \caption{Real-world examples from the Chinese platform Weibo. (a) Single-target example with one target-stance pair. (b) Multi-target example with multiple target-stance pairs.}
  \label{fig:social}
\end{figure}

However, in open social media environments, due to topic diversity and the relatedness of discussion objects, phenomena of unclear targets and multiple coexisting targets frequently commonly occur, resulting in single texts potentially containing multiple stance targets where stance labels are often associated with complex relationships between targets \cite{alturayeif2022mawqif}. Figure~\ref{fig:social} provides real examples from the Chinese platform Weibo. This work establishes a zero-shot open-world setting in which neither the target label space nor the number of targets per text is predefined, thereby requiring the model to dynamically identify all mentioned targets and assign a stance to each. Although there have been studies on target adaptation, such as an unsupervised stance detection framework combining expert mixing, domain adversarial training, and target label embeddings to achieve cross-domain prediction for unseen targets \cite{hardalov2022few}, and the Target-Stance Extraction (TSE) task which only addresses single targets by jointly modeling target identification and stance detection \cite{li2023new}, these approaches rely on target candidate labels or only support single target identification, making them difficult to adapt to multi-target and unknown target real-world application scenarios \cite{putra2022stance,sobhani2017dataset}.

% (Optional) If you REALLY want this figure to start next page, uncomment:
% \clearpage

To address these challenges, we propose a more open-ended task: Zero-Shot Stance Detection in the Wild with Dynamic Target Generation and Multi-Target Adaptation (DGTA). The task requires models to automatically identify all stance targets in text and determine the stance toward each, without relying on predefined target labels or fixed numbers of targets. We construct a high-quality Chinese multi-domain stance detection dataset, comprising 70,931 annotated samples. We design multi-dimensional evaluation metrics for target identification and stance determination, where target identification assessment includes BERTScore, BLEU, ROUGE-L, Recall and a comprehensive score, while stance determination only evaluates samples whose target identification metrics reach a threshold \cite{zhangbertscore,papineni2002bleu,lin2004rouge}. At the methodological level, we propose two strategies for fine-tuning LLMs: an integrated approach generating multiple target-stance label pairs and a two-stage method separately generating multiple targets and stance labels. We further implement a set of baseline models, including fine-tuned pre-trained models and LLMs with different prompting strategies. Experimental results demonstrate that in the DGTA task, fine-tuned LLMs significantly outperform pre-trained and prompted models, with integrated and two-stage fine-tuning strategies each showing distinct advantages.

Our main contributions are summarized as follows:
\begin{itemize}[itemsep=2pt, parsep=0pt, topsep=6pt, partopsep=0pt]
    \item We propose a new task of Zero-Shot Stance Detection in the Wild with Dynamic Target Generation and Multi-Target Adaptation (DGTA), construct the first high-quality Chinese multi-domain social media stance detection dataset, and design unified and comprehensive evaluation metrics.
    \item We explore two strategies for fine-tuning LLMs, based on integrated and two-stage frameworks, providing a powerful baseline.
    \item We conduct baseline experiments including fine-tuned pre-trained models and various prompted LLMs, with detailed comparative analysis.
\end{itemize}

\section{Dynamic Target Generation and Multi-Target Adaptation for Stance Detection}

To address the challenges of dynamic targets and complex stance relations in real-world scenarios, we define a stance detection task with dynamic target generation and multi-target adaptation. The task requires models to automatically identify multiple stance targets in text without relying on predefined targets or domains, assign stance labels to each target, and output target–stance pairs. In the following sections, we present the task definition, describe dataset construction and analysis, and introduce the evaluation criteria.

\subsection{Task Definition}

We formalize the task of \textbf{DGTA} for stance detection as follows.
Let $x$ denote an input text from a social media user. Unlike conventional stance detection, no predefined targets, topics, or domains are provided, and the number of targets in each text is not fixed.

The model is required to output a set of target–stance pairs:
\[
\mathcal{Y}(x) = \{(t_i, s_i)\}_{i=1}^{N_x},
\]
where $N_x$ is the number of targets mentioned in $x$, which may vary across texts. Each $t_i$ is a target expressed in natural language and can be either a static entity (e.g., persons, organizations, institutions) or a dynamic entity (e.g., actions, events, conceptual or abstract states). Each $s_i$ is a stance label drawn from the predefined set
\[
\mathcal{S} = \{\texttt{support}, \texttt{against}, \texttt{neutral}\}.
\]

\subsection{Chinese Social Media Persona's Stance Dataset}

\subsubsection{Data Collection and Preprocessing}

We select 240 users from diverse domains on the Weibo platform and collect their posts within the same time period. Detailed topic-user distributions are provided in \ref{sec:appendix2}. From the posts published by these 240 users, we collect a total of 125,176 textual entries. Due to the informal nature of user-generated content, we apply regular expressions and Unicode encoding techniques to remove non-standard text elements such as emojis, URLs, usernames, and special symbols, which introduce noise and reduce stance classification accuracy. During this process, all collected posts undergo strict anonymization, with user IDs anonymized. No user identity information is used in any of the experiments reported here. This anonymized ID information supports future research focusing on user-centric stance detection. Finally, after preprocessing, 107,310 posts are retained for subsequent annotation.

\subsubsection{Data Annotation and Validation}

\begin{figure}[t]
  \centering
  \includegraphics[width=\textwidth, height=0.85\textheight, keepaspectratio]{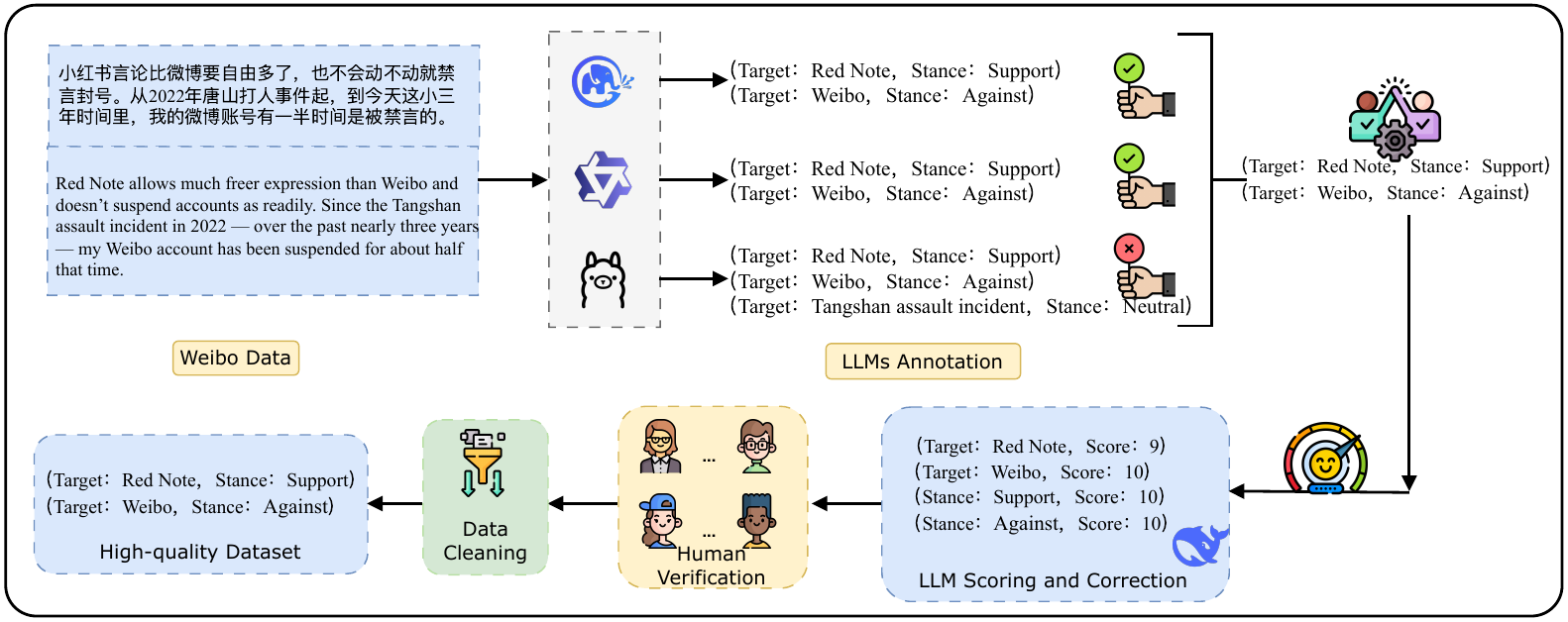}
  \caption{Workflow of dataset construction with collaboration between multiple LLMs and human verification}
  \label{fig:LLMs}
\end{figure}

To ensure the standardization and reliability of dataset annotation, we construct an annotation workflow based on the combination of collaborative annotation by multiple LLMs, score-based correction, and human verification, with the complete process illustrated in Figure~\ref{fig:LLMs}.

Specifically, we select three mainstream LLMs (GLM4-9B, Qwen2.5-7B, and Llama3-8B) to independently perform the cascaded tasks of target identification and stance determination. For the annotation results from these three models, we establish a cross-validation mechanism: in the two-stage target-stance annotation, if at least two models produce identical target entity recognition results for the same text and reach consensus on stance judgment for that target, the sample is adopted as a valid annotation; if two or more models disagree at either stage, the sample is considered invalid and removed from the dataset. After completing the first round of cross-validation, we utilize prompt instructions to guide the DeepSeek-V3 model in conducting a secondary scoring evaluation of valid annotated samples, with low-scoring samples being modified and marked for review. Subsequently, eight professional annotators verify all automatically annotated samples. More details of the full annotation workflow and inter-annotator agreement analysis are presented in \ref{sec:appendix1}. During the data cleaning phase, we eliminate 36,379 low-quality texts containing logical contradictions, semantic ambiguities, or lacking clear target references. This process ultimately results in a high-quality annotated dataset with strict cross-validation constraints.

\subsubsection{Dataset Statistics and Analysis}

After the above processing steps, the final dataset comprises 70,931 textual entries, covering both single-target and multi-target scenarios. The detailed statistics of target quantity and stance distribution are provided in Table~\ref{tab:target-stance-distribution}. Table~\ref{tab:top-targets} lists the top 10 targets by frequency, along with their mention counts and stance distributions.

% ===== Table 1 (caption moved to TOP + booktabs spacing) =====
\begin{table}[!htbp]
\centering
\caption{Target quantity and stance distribution}
\label{tab:target-stance-distribution}
\renewcommand{\arraystretch}{1.15}
\setlength{\tabcolsep}{6pt}
\begin{tabular}{lcccc}
\toprule
\multirow{2}{*}{Target} & \multirow{2}{*}{Number} & \multicolumn{3}{c}{Stance Distribution} \\
\cmidrule(lr){3-5}
 &  & Support & Against & Neutral \\
\midrule
Single & 27,148 & 12,554 & 5,232 & 9,362 \\
Dual   & 25,312 & 25,122 & 10,223 & 15,279 \\
Triple & 9,835  & 13,029 & 6,765  & 9,267 \\
Multi  & 8,636  & 22,000 & 7,809  & 17,710 \\
\midrule
\textbf{Total} & \textbf{70,931} & \textbf{72,705} & \textbf{30,029} & \textbf{51,618} \\
\bottomrule
\end{tabular}
\end{table}

% ===== Table 2 (caption moved to TOP + booktabs spacing) =====
\begin{table}[!htbp]
\centering
\caption{Top 10 targets with mention counts and stance distributions}
\label{tab:top-targets}
\renewcommand{\arraystretch}{1.15}
\setlength{\tabcolsep}{6pt}
\begin{tabular}{lcccc}
\toprule
\multirow{2}{*}{Target} & \multirow{2}{*}{Number} & \multicolumn{3}{c}{Stance Distribution} \\
\cmidrule(lr){3-5}
 &  & Support & Against & Neutral \\
\midrule
USA & 907 & 112 & 560 & 235 \\
Trump & 816 & 262 & 287 & 267 \\
Cheng Yi & 765 & 710 & 1 & 54 \\
China & 726 & 380 & 158 & 188 \\
iPhone & 567 & 150 & 127 & 290 \\
Israel & 562 & 105 & 348 & 109 \\
Wang Chuqin & 414 & 286 & 11 & 117 \\
Russia & 413 & 120 & 146 & 147 \\
Sun Yingsha & 410 & 318 & 10 & 82 \\
Huawei & 350 & 256 & 9 & 85 \\
\bottomrule
\end{tabular}
\end{table}

\subsection{Evaluation Criteria}

Due to the diversity and uncertainty in both expression and quantity of dynamically generated targets in this task, traditional evaluation metrics fail to adequately reflect model performance. To comprehensively assess model performance in this task, we design more targeted and comprehensive evaluation criteria.

\subsubsection{Target Identification Evaluation Criteria}

For the open-ended characteristics of the target identification phase, we propose a multi-dimensional evaluation approach that integrates semantic similarity, surface form matching, and quantity alignment to construct a comprehensive target identification score (C-Score). This metric comprises BERTScore, BLEU, ROUGE-L, and the Recall of target quantity.

\begin{small}
\begin{equation}
C\text{-Score} = (\alpha \times \text{BERTScore} + \beta \times \text{BLEU} + \gamma \times \text{ROUGE-L}) \times \text{Recall}
\label{eq:cscore}
\end{equation}
\end{small}

Where $\alpha$, $\beta$, and $\gamma$ control the weighting proportions of the three metrics. Considering the semantic and structural differences between predicted targets and reference targets, we first align the metrics. Experiments show that setting $\alpha=0.6$, $\beta=0.2$, and $\gamma=0.2$ emphasizes semantic consistency while balancing lexical and structural matching.

\subsubsection{Stance Detection Evaluation Criteria}

Considering that stance classification evaluation is only meaningful when based on accurate target identification, we first establish a threshold-driven mechanism for determining target correctness: through experimental analysis, we set thresholds of 0.7, 0.2, 0.4, 0.8, and 0.3 for BERTScore, BLEU, ROUGE-L, Recall, and C-Score, respectively. Samples exceeding these thresholds are deemed to have correct target identification. On this foundation, we employ the classic metrics of Precision, Recall, and F1 score.

\section{Fine-tuning LLMs}

\begin{figure}[!htbp]
  \centering
  \includegraphics[width=1.0\linewidth]{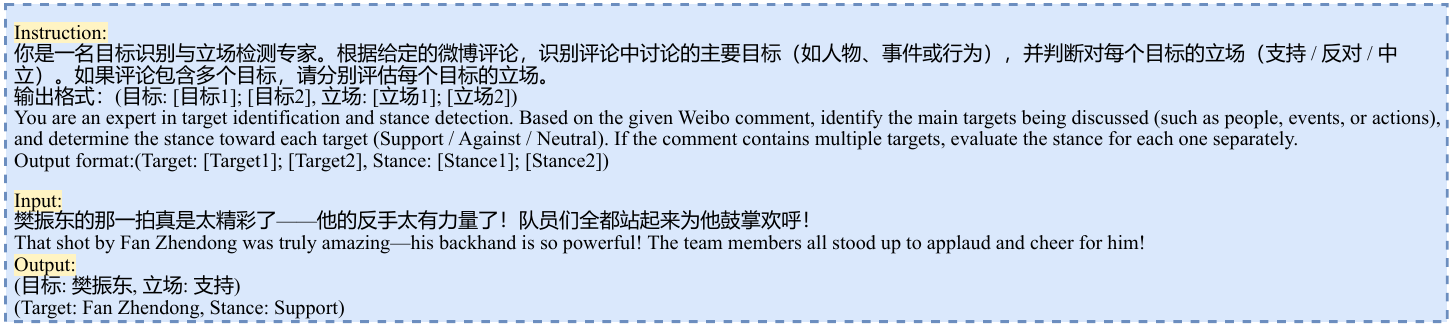}
  \caption{Prompt template and example for the integrated fine-tuning strategy}
  \label{fig:integrated}
\end{figure}

We fine-tune LLMs to provide powerful baselines for this task. We propose two fine-tuning strategies—integrated and two-stage, and use each strategy to construct instruction fine-tuning data. All fine-tuning is conducted using LoRA \cite{hu2022lora}. The integrated fine-tuning strategy adopts an end-to-end approach, modeling the "target identification + stance detection" task as an instruction-driven sequence generation process. Model input consists of task instructions in natural language concatenated with the original text (as shown in Figure~\ref{fig:integrated}), explicitly prompting the task intent, guiding the model to simultaneously complete target extraction and stance classification, ultimately outputting (multiple) target-stance pairs, achieving task coordination.

\begin{figure}[!htbp]
  \centering
  \includegraphics[width=\linewidth,height=0.30\textheight,keepaspectratio]{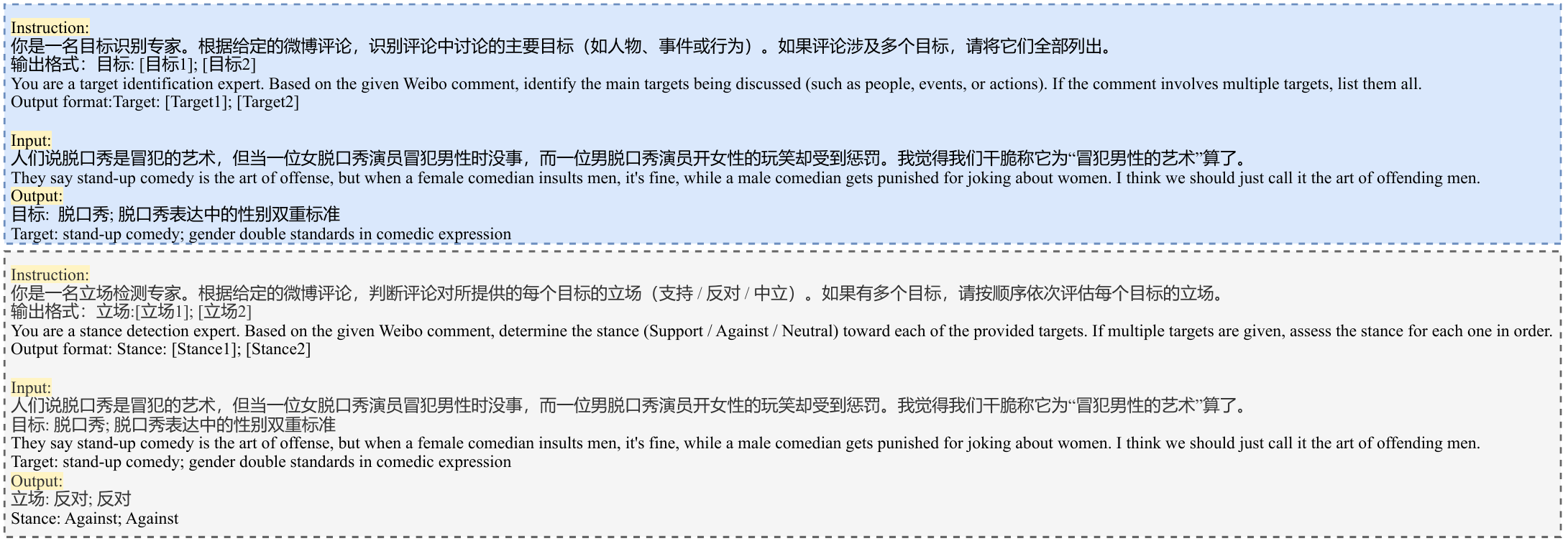}
  \caption{Prompt template and example for the two-stage fine-tuning strategy}
  \label{fig:two-stage}
\end{figure}

The two-stage fine-tuning strategy decouples target identification and stance determination into two independent subtasks, each undergoing separate instruction fine-tuning. In the first stage, the model receives input text with task instructions (as shown in Figure~\ref{fig:two-stage}), focusing exclusively on extracting potential targets from the text. In the second stage, the identified targets and original text serve as input (as shown in Figure~\ref{fig:two-stage}), accompanied by stance determination instructions, guiding the model to classify stance for specific targets. Through independent fine-tuning, models can focus on a single task. It is worth noting that we use different models for our two-stage fine-tuning approach, rather than the same model.

% NOTE: remove manual \vspace{-10pt}; rely on \floatsep/\textfloatsep for stable spacing
\vspace{-4pt}

\section{Experiments and Analysis}
\subsection{Experimental Setups}
\subsubsection{Dataset}

We divide our constructed dataset into training, validation, and test sets in an 8:1:1 ratio for fine-tuning and baseline evaluation experiments. Considering the high computational resources required for evaluating the full dataset, we construct the stratified test subset using fixed seeds $\{13,97,233,521,907,2029,4099\}$ (i.e., $K=7$ draws of $1{,}000$ instances), ensuring the joint distribution over target and stance matches the full test partition. We then evaluate all models on each draw and report the average performance and standard deviation, together with 95\% confidence intervals.

\subsubsection{Evaluation Metric Parameter Setting}

All evaluation parameters are determined through systematic analysis on the development set. For the C-Score, we selected the weight parameters $\alpha = 0.6$, $\beta = 0.2$, $\gamma = 0.2$ via grid search to maximize correlation with expert annotations (Fleiss's~$\kappa = 0.87$, $n = 500$) on the development set. Target correctness thresholds are established through ROC and precision-recall analysis on 500 manually verified samples (see \ref{sec:appendix3}), yielding: BERTScore $\geq 0.7$, BLEU $\geq 0.2$, ROUGE-L $\geq 0.4$, Recall $\geq 0.8$, and C-Score $\geq 0.3$, achieving 89.2\% agreement with human judgments. Only target-stance pairs meeting all criteria are evaluated for stance classification.

\subsubsection{Comparison Models}

We compare three categories of models: fine-tuned pre-trained models, instruction-prompted LLMs, and fine-tuned LLMs. For the fine-tuned pre-trained models, we employ fine-tuned mT5 for target identification and fine-tuned BERT for stance detection \cite{xue2021mt5,devlin2019bert}. In addition, we include two strong supervised baselines: (i) a pipeline that first performs target mention extraction with RoBERTa-CRF and then applies a target-conditioned RoBERTa-large classifier for stance, and (ii) an end-to-end mT5 model that directly generates linearized \((\textit{target};\,\textit{stance})\) tuples via constrained decoding. For the instruction-prompted LLMs, we experiment with current mainstream models including DeepSeek-V3, GLM4-9B, GPT-4o, and Llama3-8B using instruction prompting \cite{liu2024deepseek,glm2024chatglm,hurst2024gpt,grattafiori2024llama}. For the fine-tuned LLMs, we fine-tune Qwen2.5-7B-Instruct and DeepSeek-R1-Distill-Qwen-7B using an integrated approach, and also independently fine-tune Qwen2.5-7B-Instruct for target identification and DeepSeek-R1-Distill-Qwen-7B for stance detection in a two-stage process \cite{Qwen2024Qwen2,guo2025deepseek}.

\subsection{Experimental Results and Analysis}

% ===== Main results table (caption moved to TOP; keep resizebox but loosen row/col spacing) =====
\begin{table*}[t]
\caption{Overall experimental results on the DGTA task (Unit: \%, best results are in bold. † indicates integrated fine-tuning; ‡ indicates the target identification stage in two-stage fine-tuning; § indicates the stance determination stage in two-stage fine-tuning. DeepSeek-R1-Distill-Qwen-7B is abbreviated as DeepSeek-R1-Qwen. BERTScore is abbreviated as BERT-S, and ROUGE-L is abbreviated as ROUGE.)}
\label{tab:main-test}
\centering

\renewcommand{\arraystretch}{1.18} % 行距：1.15~1.25 自行微调
\setlength{\tabcolsep}{4.5pt}     % 列间距：4~6pt 自行微调

\resizebox{\textwidth}{!}{%
\begin{tabular}{c|ccccc|ccc}
\hline
\multirow{2}{*}{Model} & \multicolumn{5}{c|}{Target Identification} & \multicolumn{3}{c}{Stance Detection} \\
\cline{2-9}
 & BERT-S & BLEU & ROUGE & Recall & C-Score & P & R & F1 \\
\hline
mT5$^{\ddagger}$ & 84.29 & 28.82 & 65.71 & 86.59 & 60.16 & - & - & - \\
Bert$^{\mathsection}$ & - & - & - & - & - & 67.89 & 67.11 & 67.51 \\
mT5-e2e$^{\dagger}$ & 83.42 & 29.67 & 65.88 & 89.73 & 62.06 & 66.12 & 63.41 & 64.74 \\
RoBERTa-CRF$^{\ddagger}$ & 84.07 & 30.43 & 66.18 & 88.97 & 62.07 & - & - & - \\
RoBERTa-large $^{\mathsection}$ & - & - & - & - & - & 65.74 & 59.01 & 62.19 \\
\hline
Qwen2.5-7B & 82.47 & 28.26 & 63.69 & 91.16 & 61.87 & 64.22 & 67.54 & 65.05 \\
DeepSeek-V3 & 76.87 & 25.65 & 50.38 & 92.05 & 56.45 & 69.25 & 72.60 & 70.64 \\
GLM4-9B & 77.98 & 24.16 & 51.99 & 94.14 & 58.38 & 68.50 & 69.20 & 66.90 \\
GPT-4o & 73.72 & 21.51 & 43.99 & 94.34 & 54.09 & 74.22 & 74.75 & 74.45 \\
Llama3-8B & 77.69 & 27.94 & 56.98 & 85.90 & 54.63 & 58.45 & 65.03 & 59.52 \\
\hline
Qwen2.5-7B$^{\dagger}$ & \textbf{85.09} & 31.12 & \textbf{67.14} & 94.16 & 66.58 & 65.31 & 65.33 & 64.16 \\
DeepSeek-R1-Qwen$^{\dagger}$ & 84.94 & 30.96 & 66.99 & 94.62 & 66.76 & \textbf{87.46} & \textbf{77.08} & \textbf{79.26} \\
Qwen2.5-7B$^{\ddagger}$ & 84.69 & \textbf{31.64} & 66.17 & \textbf{95.19} & \textbf{66.99} & - & - & - \\
DeepSeek-R1-Qwen$^{\mathsection}$ & - & - & - & - & - & 83.25 & 74.52 & 75.37 \\
\hline
\end{tabular}%
}
\end{table*}

The experimental results are shown in Table~\ref{tab:main-test}, and we can find that:

\begin{itemize}
    \item Fine-tuned LLMs significantly outperform both fine-tuned pre-trained models and instruction-prompted LLMs on the DGTA task. Both integrated and two-stage fine-tuned LLMs achieve comprehensive scores exceeding 66\% in target identification, with Qwen2.5-7B demonstrating optimal performance (66.99\%). In target identification tasks, both fine-tuned and pre-trained models exhibit BERTScore metrics above 84\%, indicating that fine-tuning enhances target semantic comprehension capabilities. For stance detection, fine-tuned DeepSeek-R1 models achieve F1 scores (79.26\% and 75.37\%) that surpass Llama3-8B by over 15\%, demonstrating that fine-tuning substantially improves stance reasoning abilities in complex semantic contexts.
    \item Integrated and two-stage strategies each have advantages in target identification and stance detection subtasks. For target identification, two-stage fine-tuning enables greater focus, with Qwen2.5-7B achieving the optimal score of 66.99\%. For stance detection, integrated fine-tuning exhibits superior performance, with DeepSeek-R1-Distill-Qwen-7B outperforming two-stage models across all evaluation metrics, likely due to its ability to simultaneously model inter-target relationships and stance associations.
    \item Models with reasoning capabilities generally perform better than those without. Between the two integrated fine-tuned models—Qwen2.5-7B and DeepSeek-R1-Distill-Qwen-7B—the latter underwent reasoning distillation. Comparison reveals that the reasoning-capable DeepSeek-R1 model achieves a comprehensive score of 66.76\% in target identification and an F1 score of 79.26\% in stance detection, outperforming the Qwen2.5-7B model overall. This indicates that reasoning capabilities contribute to more precise target identification and stance determination in complex scenarios.
\end{itemize}

\subsection{Dynamic Target Difference Analysis}

% ===== Target-diff table (caption TOP + booktabs) =====
\begin{table}[htbp]
\centering
\caption{Target identification performance across different target quantities. (Unit: \%. $^\dagger$ indicates integrated fine-tuning. Best results are in bold.)}
\label{tab:target-diff}
\renewcommand{\arraystretch}{1.15}
\setlength{\tabcolsep}{5pt}
\begin{tabular}{llccccc}
\toprule
Model & Target & BERT-S & BLEU & ROUGE & Recall & C-Score \\
\midrule
\multirow{4}{*}{Qwen2.5-7B$^\dagger$} 
  & Single & 83.90 & 29.14 & 64.77 & \textbf{99.99} & \textbf{69.11} \\
  & Dual   & \textbf{86.89} & \textbf{33.46} & \textbf{71.22} & 94.26 & 68.98 \\
  & Triple & 85.78 & 33.03 & 68.65 & 85.45 & 61.45 \\
  & Multi  & 83.64 & 30.11 & 62.08 & 75.17 & 53.58 \\
\midrule
\multirow{4}{*}{All Models Avg}
  & Single & 82.04 & 28.79 & 61.55 & \textbf{99.36} & \textbf{67.28} \\
  & Dual   & \textbf{82.47} & \textbf{29.47} & \textbf{62.72} & 93.59 & 63.76 \\
  & Triple & 80.52 & 27.89 & 58.01 & 80.43 & 53.02 \\
  & Multi  & 79.29 & 26.80 & 55.34 & 67.67 & 45.42 \\
\bottomrule
\end{tabular}
\end{table}

We conduct a comprehensive comparison of model performance across different target quantities, as shown in Table~\ref{tab:target-diff}. Samples are categorized into single-target, dual-target, triple-target, and multi-target groups (more than three targets).

Dual-target samples yield the highest scores on semantic-oriented metrics (BERTScore, BLEU, and ROUGE-L) for both Qwen2.5-7B$^\dagger$ and the model average. This suggests that dual-target texts offer favorable conditions for accurate semantic alignment. One possible explanation is that such texts often involve contrastive or comparative structures, which help clarify entity boundaries and highlight semantic distinctions. Additionally, dual-target texts tend to balance information richness and complexity, avoiding the sparsity of single-target texts and the semantic complexity of multi-target cases.

Single-target samples achieve the highest recall (99.36\% on average; 99.99\% for Qwen2.5-7B$^\dagger$), likely due to their focus on a single, unambiguous entity, which enables exhaustive extraction. However, their relatively lower scores on semantic metrics indicate that while models can detect targets reliably, they may struggle to fully capture nuanced or implicit references, especially in shorter or contextually underspecified utterances.

As the number of targets increases, all metrics exhibit a downward trend. Triple- and multi-target samples show notable performance degradation, with the lowest C-Scores observed in the multi-target category (e.g., 45.42\% on average; 53.58\% for Qwen2.5-7B$^\dagger$). This decline can be attributed to increased semantic complexity, more frequent overlapping or nested references, and the model's tendency to conflate multiple entities into a single representation—negatively impacting both recall and semantic precision.

\subsection{Impact of Chain-of-Thought on Prompted LLMs}

We investigate whether introducing chain-of-thought (CoT) improves LLM performance on the DGTA task. The results are presented in Table~\ref{tab:cot}.

% ===== CoT table (caption TOP + booktabs; keep resizebox but loosen) =====
\begin{table*}[htbp]
\centering
\caption{Experimental results with chain-of-thought incorporated in the prompt (Unit: \%)}
\label{tab:cot}

\renewcommand{\arraystretch}{1.25} % 1.15~1.35 之间都可以试
\setlength{\tabcolsep}{4pt}

\begin{tabular}{c|ccccc|ccc}
\hline
\multirow{2}{*}{Model} & \multicolumn{5}{c|}{Target Identification} & \multicolumn{3}{c}{Stance Detection} \\
\cline{2-9}
 & BERT-S & BLEU & ROUGE & Recall & C-Score & P & R & F1 \\
\hline
Qwen2.5-7B        & 82.47 & 28.26 & 63.69 & 91.16 & 61.87 & 64.22 & 67.54 & 65.05 \\
DeepSeek-V3       & 76.87 & 25.65 & 50.38 & 92.05 & 56.45 & 69.25 & \textbf{72.60} & 70.64 \\
GLM4-9B           & 77.98 & 24.16 & 51.99 & 94.14 & 58.38 & 68.50 & 69.20 & 66.90 \\
\hline
CoT-Qwen2.5-7B    & 84.97 & 31.02 & 67.77 & \textbf{94.23} & \textbf{66.70} & 69.07 & 70.84 & 69.25 \\
CoT-DeepSeek-V3   & 82.92 & \textbf{32.33} & 62.62 & 94.18 & 64.75 & \textbf{73.68} & 71.42 & \textbf{71.06} \\
CoT-GLM4-9B       & \textbf{85.56} & 31.46 & \textbf{68.38} & 92.03 & 65.63 & 69.38 & 69.22 & 67.91 \\
\hline
\end{tabular}
\end{table*}

After introducing CoT, all LLMs show significant improvements in both target identification and stance determination. GLM4-9B's C-Score increases by 7 percentage points, indicating that step-by-step reasoning more effectively guides the model to capture key targets. Qwen2.5-7B's F1 score improves by 4 percentage points, as the reasoning chain encourages the model to analyze systematically, reducing inferential leaps and incorrect judgments, thereby significantly enhancing stance classification performance.

\subsection{Target Significance Difference Analysis}

% ===== Explicit/Implicit table (caption TOP + booktabs; avoid ultra-tight) =====
\begin{table}[htbp]
\centering
\caption{Performance comparison on explicit and implicit targets. (Unit: \%. † indicates integrated fine-tuning; DeepSeek-R1-Distill-Qwen-7B is abbreviated as DeepSeek-R1-Qwen. Explicit targets cover 80.51\% of the data, while implicit targets cover 19.48\%.)}
\label{tab:explicit-implicit}

\renewcommand{\arraystretch}{1.25} % 1.20~1.35 自己微调
\setlength{\tabcolsep}{4pt}

\resizebox{\linewidth}{!}{%
\begin{tabular}{c|c|ccccc|ccc}
\hline
\multirow{2}{*}{Target} & \multirow{2}{*}{Model} & \multicolumn{5}{c|}{Target Identification} & \multicolumn{3}{c}{Stance Detection} \\
\cline{3-10}
 &  & BERT-S & BLEU & ROUGE & Recall & C-Score & P & R & F1 \\
\hline
\multirow{2}{*}{\makecell{Explicit \\ (80.51\%)}} 
    & DeepSeek-R1-Qwen$^{\dagger}$ & 87.57 & 35.06 & 73.14 & 94.15 & 70.17 & \textbf{69.40} & \textbf{72.95} & \textbf{70.88} \\
    & Qwen2.5-7B$^{\dagger}$       & \textbf{87.79} & \textbf{35.25} & \textbf{73.39} & 93.85 & \textbf{70.24} & 65.34 & 65.54 & 64.09 \\
\hline
\multirow{2}{*}{\makecell{Implicit \\ (19.48\%)}} 
    & DeepSeek-R1-Qwen$^{\dagger}$ & 73.35 & 12.88 & 39.93 & \textbf{96.63} & 52.34 & 63.21 & 64.50 & 63.54 \\
    & Qwen2.5-7B$^{\dagger}$       & 73.22 & 12.93 & 39.56 & 95.51 & 51.62 & 65.15 & 63.29 & 63.90 \\
\hline
\end{tabular}%
}
\end{table}

Considering different topic backgrounds and expression styles, targets in texts exhibit varying degrees of salience. We employ DeepSeek-V3 to classify annotated targets in the extracted test set as either "explicit" or "implicit", where the former refers to directly mentioned specific entities and the latter to abstract concepts requiring semantic understanding.

Based on experimental results, we select the high-performing integrated fine-tuned models DeepSeek-R1-Distill-Qwen-7B and Qwen2.5-7B for statistical analysis of target salience. Table~\ref{tab:explicit-implicit} analysis reveals.

In target identification tasks, models perform significantly better when processing explicit targets compared to implicit ones. For instance, Qwen2.5-7B scores notably higher across multiple metrics, indicating that explicit targets have clearer semantic boundaries, facilitating extraction and matching.

In target identification tasks, implicit targets demonstrate more prominent performance in terms of recall. DeepSeek-R1-Distill-Qwen-7B achieves a recall rate of 96.63\% for implicit targets. Due to the abstract nature of implicit targets, models tend to generate multiple related expressions for coverage, enhancing recall but potentially reducing precision.

In stance detection tasks, explicit targets similarly demonstrate superior detection performance. Explicit targets help models more accurately grasp user attitudes, improving F1 scores, while implicit targets increase judgment difficulty due to semantic ambiguity. Models equipped with reasoning capabilities can further enhance performance in these scenarios.

% NOTE: remove manual \vspace{-10pt}; rely on caption/float spacing globally
\begin{figure}[htbp]
  \centering
  \includegraphics[width=\linewidth,height=0.45\textheight,keepaspectratio,
  trim=0 0 0 0,clip]{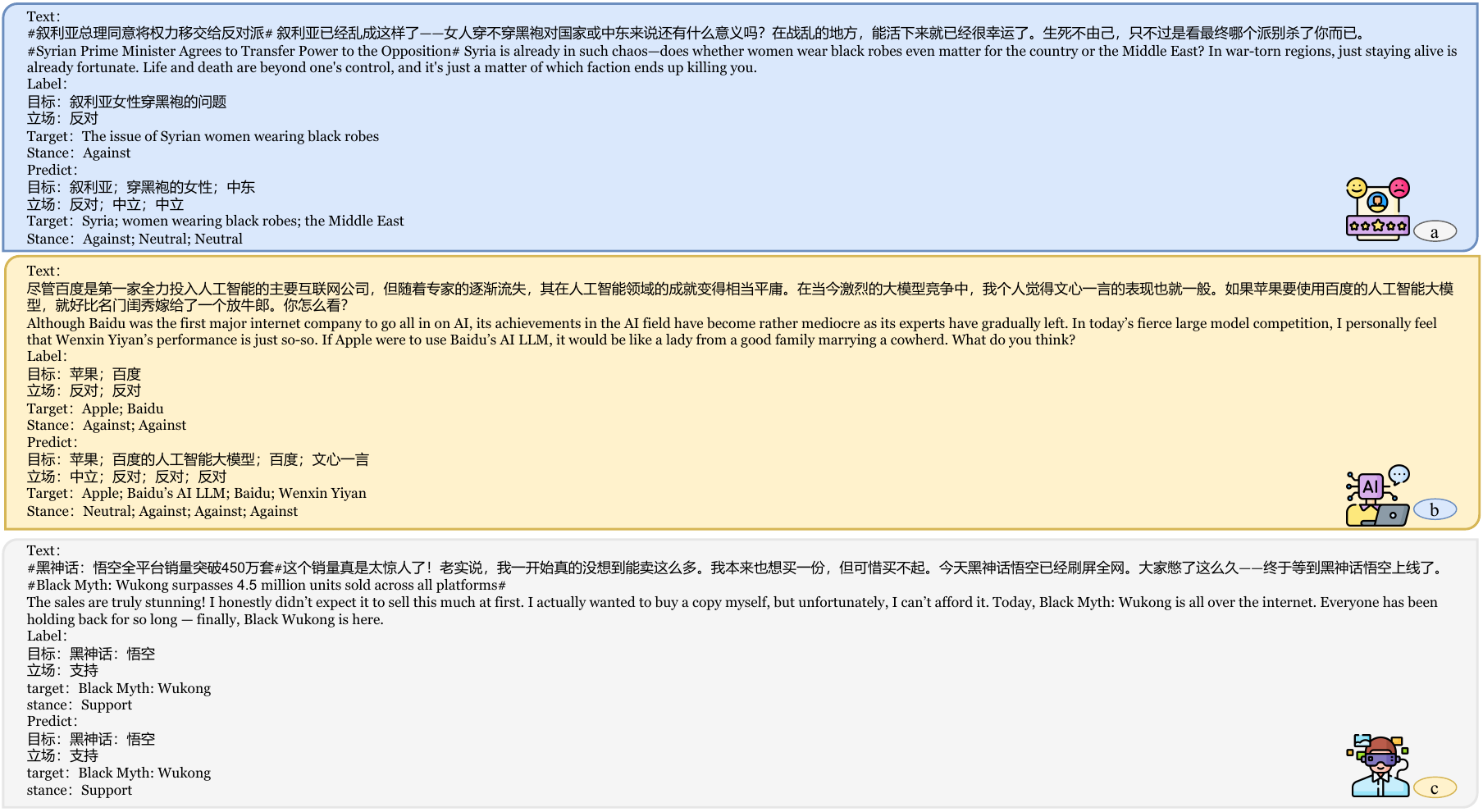}
  \caption{Three representative cases}
  \label{fig:case}
\end{figure}

\subsection{Case Study}

We randomly sample cases from the integrated fine-tuned model DeepSeek-R1-Distill-Qwen-7B's prediction results for case analysis (Figure~\ref{fig:case}).

Target identification performs well overall, but semantic fragmentation and stance judgment biases remain. In case (a), "the issue of Syrian women wearing black robes" is decomposed into "Syria", "women wearing black robes" and "Middle East" resulting in a loss of semantic integrity. Simultaneously, the model fails to identify the implicit critical attitude in "women wearing black robes" incorrectly judging it as neutral, reflecting its insufficient ability to reason about irony or implicit semantics. 

Inconsistent target granularity and insufficient understanding of sarcastic expressions are observed. In case (b), although the model can extract multiple targets, it exhibits problems with mixed usage of different expressions for the same object, such as "Baidu's AI LLM" and "Wenxin Yiyan" both referring to "Baidu". Additionally, the stance judgment toward "Apple" as neutral fails to identify the metaphorical expression "a lady from a good family marrying a cowherd" reflecting the model's difficulty in recognizing stance under non-straightforward expressions like sarcasm and metaphor.

The model demonstrates optimal performance in scenarios with clearly defined targets and explicitly expressed stances. Case (c) revolves around the single target "Black Myth: Wukong" with direct textual expressions and distinct emotions, such as "stunning" and "holding back for so long" clearly conveying a positive stance. The model accurately identifies the target and correctly judges the stance, indicating high predictability in such samples.

\section{Related Work}

\subsection{Traditional Stance Detection}

Traditional stance detection methods have evolved from manual feature engineering to contextualized pre-trained models \cite{glandt2021stance,zxb2024Survey}. Zarrella and Marsh integrate grammatical and syntactic information into RNNs and learns vector representations of input text, effectively enhancing stance detection performance on Twitter texts \cite{2016MITRE}. Du et al. introduce attention mechanisms into LSTM, proposing a target-specific enhanced attention model \cite{du2017stance} . WS-BERT substantially improves performance in target-specific \cite{he2022infusing}, cross-target, and zero/few-shot scenarios by integrating Wikipedia knowledge to enrich target representations. The GDA-CL model generates high-quality synthetic samples in embedding space through generative adversarial networks (GAN) and hybrid contrastive learning \cite{li2022generative}, using GPT-2 as the generator, RoBERTa as the discriminator, and BERT as the classifier within the GAN framework \cite{goodfellow2014generative}, supplemented with a multilayer perceptron for contrastive learning, significantly improving zero-shot stance detection performance on unseen targets. Stance Reasoner aims to leverage explicit reasoning about background knowledge to guide models in inferring target stances \cite{taranukhin2024stance}. LKI-BART introduces LLM knowledge to establish connections between text and unseen targets, achieving optimal performance on VAST and P-Stance datasets \cite{zhang2024llm,allaway2020zero,li2021p}. Zhang et al. are advancing LLM stance detection with prompt learning, showing labels and brief context boost accuracy \cite{yuanshuo2024}. MSME is a multi-expert zero-shot stance framework that fuses retrieved knowledge and pragmatic cues for robust prediction \cite{zhang2025msme}. However, these studies all rely on predefined targets, cannot adapt to real-world scenarios where targets are implicit or unknown, and fail to address the problem of dynamic target generation.

\subsection{Target-Adaptive Stance Detection}

Recent studies have begun to focus on target-adaptive stance detection. TATA leverages contrastive learning to extract topic-agnostic and topic-aware embeddings from unlabeled news texts and applies them to downstream stance detection tasks \cite{hanley2023tata}. TAPD enhances cross-target few-shot stance detection through target-aware prompt adaptation and multi-prompt distillation techniques \cite{wang2024target,wen2023zero}, mapping stance labels to continuous vectors. For cross-target domain adaptation, the target-aware domain adaptation method extracts key shared features through feature disentanglement and automatically identifies target relationships \cite{deng2022domain}. Stanceformer introduces target-aware attention mechanisms \cite{garg2024stanceformer}. OpenStance
defines the open-domain zero-shot stance detection task \cite{xu2022openstance}, addressing stance detection without domain restrictions or specific topic focus. Wu proposes a novel multi-source adaptive target detection method for Target-Related Knowledge Preservation \cite{wu2022target}. For the new task of cross-lingual cross-target stance detection, a dual-teacher knowledge distillation framework CCSD is designed \cite{zhang2023cross}, utilizing cross-lingual and cross-target teachers to guide student model learning from source languages. ZeroStance leverages ChatGPT to synthesize multi-domain data for zero-shot stance detection, enhancing model generalization to unseen targets across diverse domains \cite{zhao2024zerostance}. Lan et al. propose a new target-adaptive, interpretable stance detection task \cite{lan2025targetadaptive}. Although these works have significantly advanced target-adaptive stance detection, they still rely on predefined target lists or domain-specific data and only address single-target adaptation, limiting model adaptability for stance detection in real scenarios with undefined, multiple targets requiring dynamic generation and adaptation.

\section{Conclusion}

We introduce a zero-shot stance detection task in the wild, incorporating dynamic target generation and multi-target adaptation, to handle the complexity and diversity of stance expressions in social media. We construct a large-scale Chinese dataset covering multiple social scenarios and develop an evaluation framework that jointly assesses target identification and stance classification. Two fine-tuning strategies for LLMs are implemented and compared with pre-trained models and models under different prompting methods. Experimental results indicate that fine-tuned models achieve higher accuracy in target extraction and more consistent stance classification across diverse contexts, establishing a solid baseline for future research in this area.

\bibliographystyle{unsrtnat}
\bibliography{references}

\newpage
\appendix
\section*{Appendix}

\section{Topic--User Distribution}
\label{sec:appendix2}

\begin{figure}[htbp]
  \centering
  \includegraphics[width=0.8\linewidth]{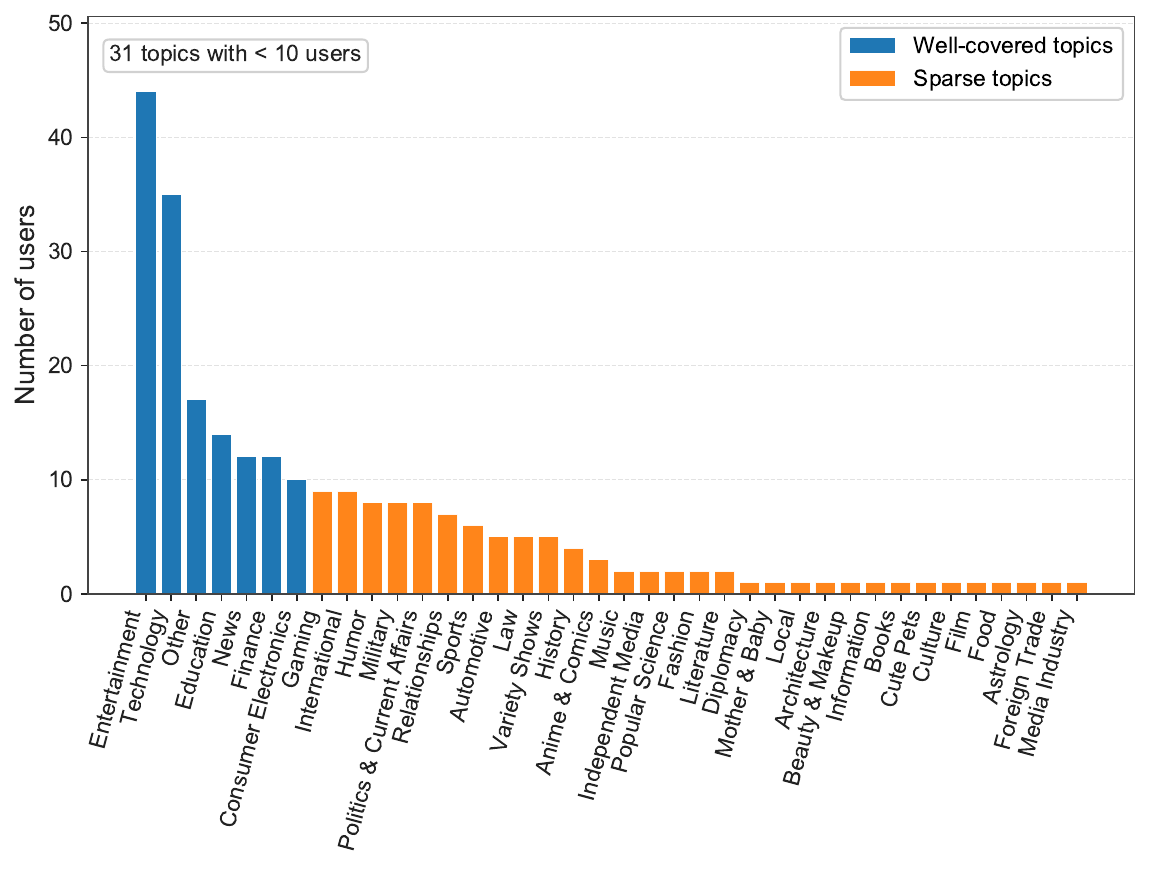}
  \caption{Topic coverage of the 240 sampled Weibo users. Each bar shows
  the number of users per fine-grained topic; blue bars mark topics with
  at least 10 users and orange bars mark sparse topics with fewer than
  10 users.}
  \label{fig:user-coverage}
\end{figure}

\begin{figure}[htbp]
  \centering
  \includegraphics[width=0.5\linewidth]{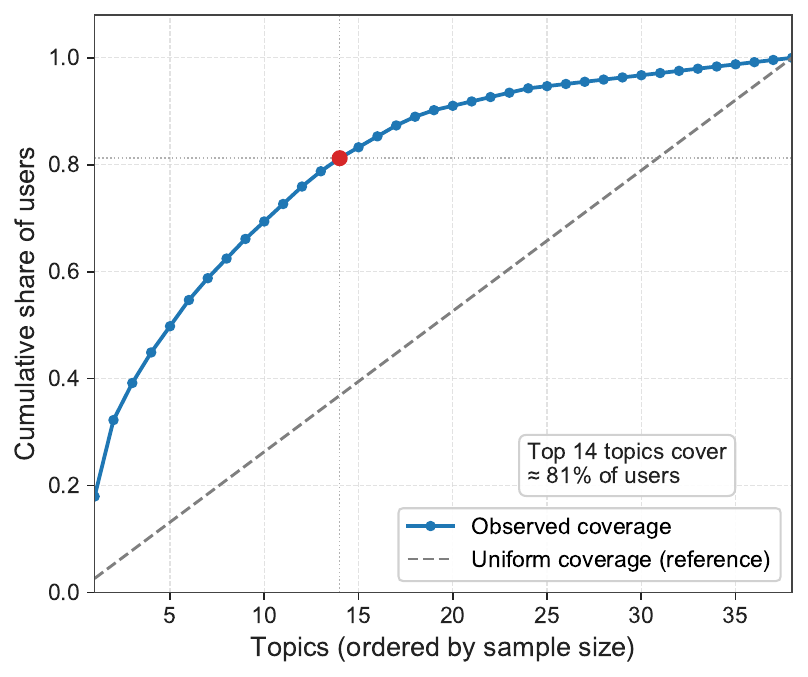}
  \caption{Cumulative share of users as topics are added in order of
  decreasing size. The blue curve shows the observed coverage and the
  grey dashed line a uniform reference. The red point indicates that
  the top 14 topics already cover about 81\% of users.}
  \label{fig:user}
\end{figure}

Figures~\ref{fig:user-coverage} and~\ref{fig:user} summarize the topic coverage of the 240 sampled Weibo users. The corpus spans 36 fine-grained topics across entertainment, technology, finance, law, education and other domains, reducing single-domain bias. At the same time, the distribution is long-tailed: 31 topics contain fewer than 10 users, and the 14 most frequent topics cover about 81\% of all users. Thus the dataset is most reliable for common topics, and generalization to rare topics, unseen users or other settings may be limited and warrants further validation. All statistics are computed on anonymized,de-identified aggregates.

\section{Annotation Quality and Inter-Annotator Agreement}
\label{sec:appendix1}

\subsection{Human Annotation and Inter-Annotator Agreement (IAA)}

Eight professional annotators verify all labels under a unified guideline. On the full dataset (\(N = 107{,}310\)), Fleiss’s~\(\kappa\) is \(0.82\) (95\% CI: [0.80, 0.84]), indicating \textit{almost perfect} agreement~\cite{nichols2010putting}. Agreement is \(0.85\) (95\% CI: [0.83, 0.87]) for single-target and \(0.78\) (95\% CI: [0.76, 0.80]) for multi-target instances; Krippendorff’s~\(\alpha = 0.81\) and exact-match agreement \(>89\%\) corroborate this reliability.

\subsection{Disagreement and Adjudication Protocol}

Each sample is independently reviewed. When annotators disagree on the target span or stance polarity, the instance enters a three-annotator adjudication process involving a senior reviewer~\cite{li2021p-stance}. Final decisions are made by majority vote or brief consensus discussion. Instances with irreconcilable ambiguity (e.g., logical contradictions or unclear target reference) are excluded during final cleaning. This procedure prevents difficult cases from artificially lowering IAA while maintaining a high-quality dataset for modeling and evaluation.

\FloatBarrier

\section{ROC and Precision--Recall Curves for Target Metrics}
\label{sec:appendix3}

\begin{figure}[htbp]
  \centering
  \scriptsize

  % -------- Row 1: ROC --------
  \begin{minipage}[b]{0.19\textwidth}
    \centering
    \includegraphics[width=\linewidth]{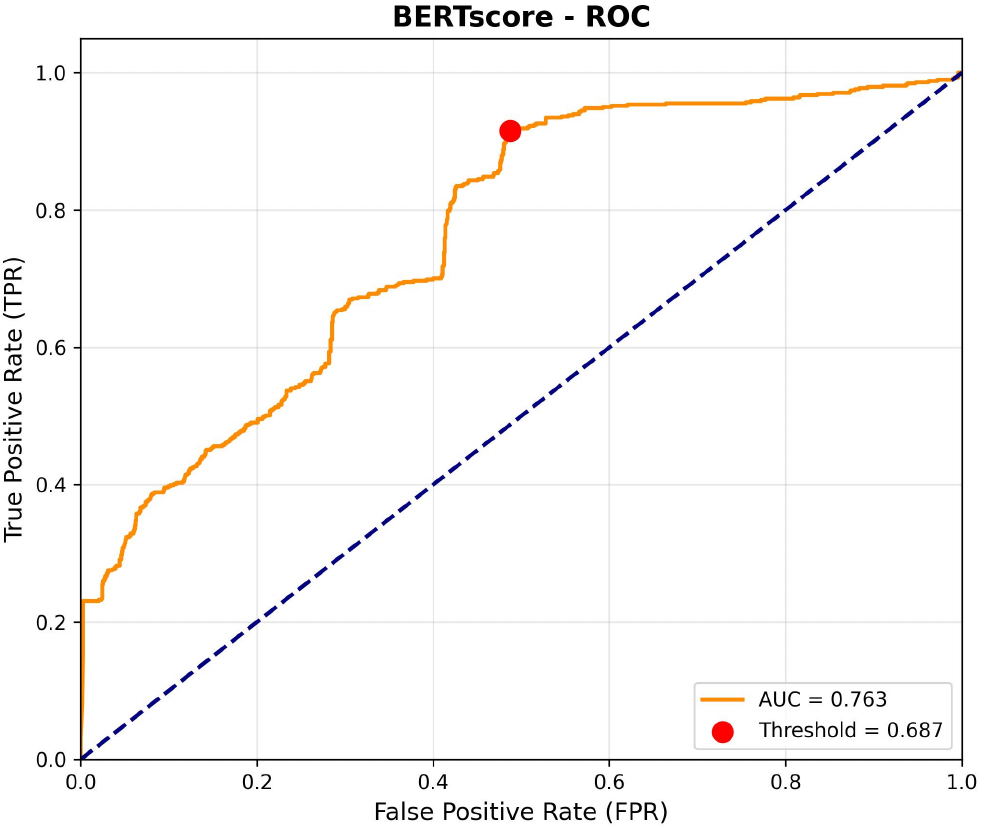}\\[-2pt]
    (a) BERTScore(ROC)
  \end{minipage}\hfill
  \begin{minipage}[b]{0.19\textwidth}
    \centering
    \includegraphics[width=\linewidth]{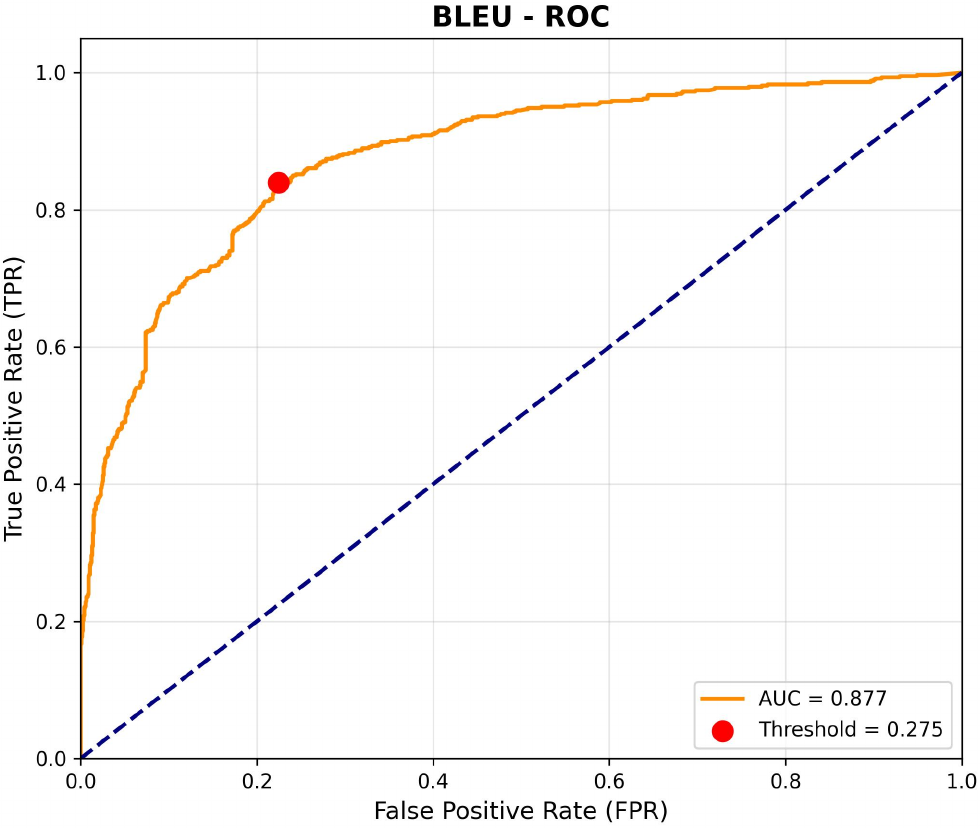}\\[-2pt]
    (b) BLEU(ROC)
  \end{minipage}\hfill
  \begin{minipage}[b]{0.19\textwidth}
    \centering
    \includegraphics[width=\linewidth]{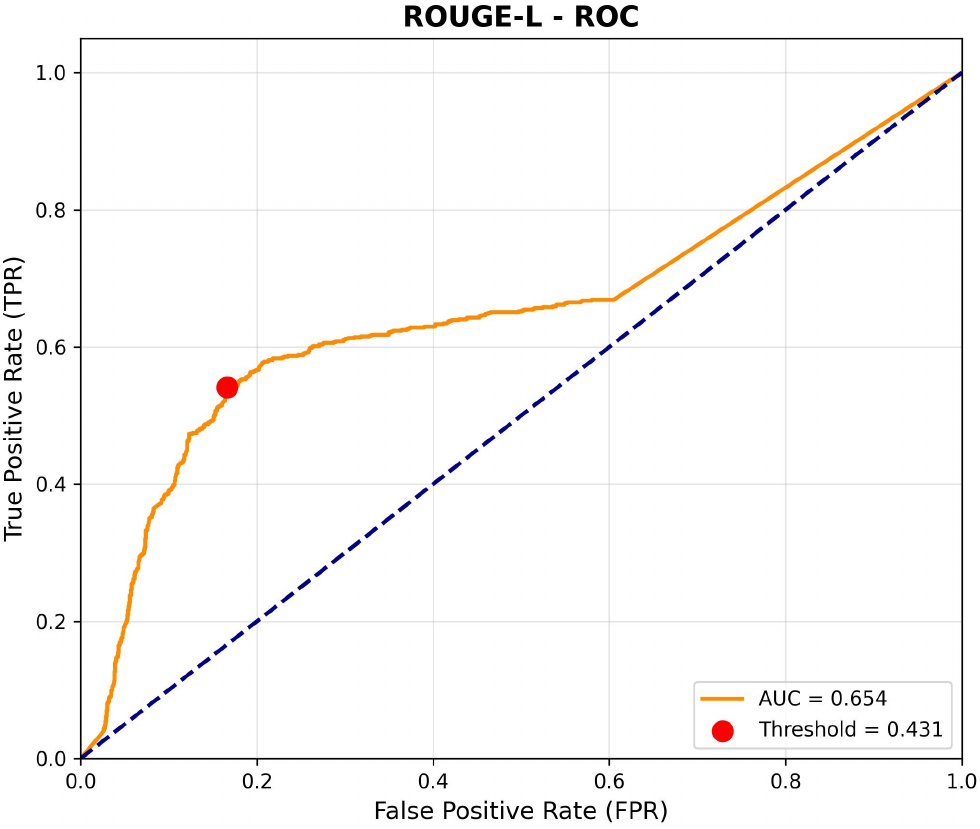}\\[-2pt]
    (c) ROUGE-L(ROC)
  \end{minipage}\hfill
  \begin{minipage}[b]{0.19\textwidth}
    \centering
    \includegraphics[width=\linewidth]{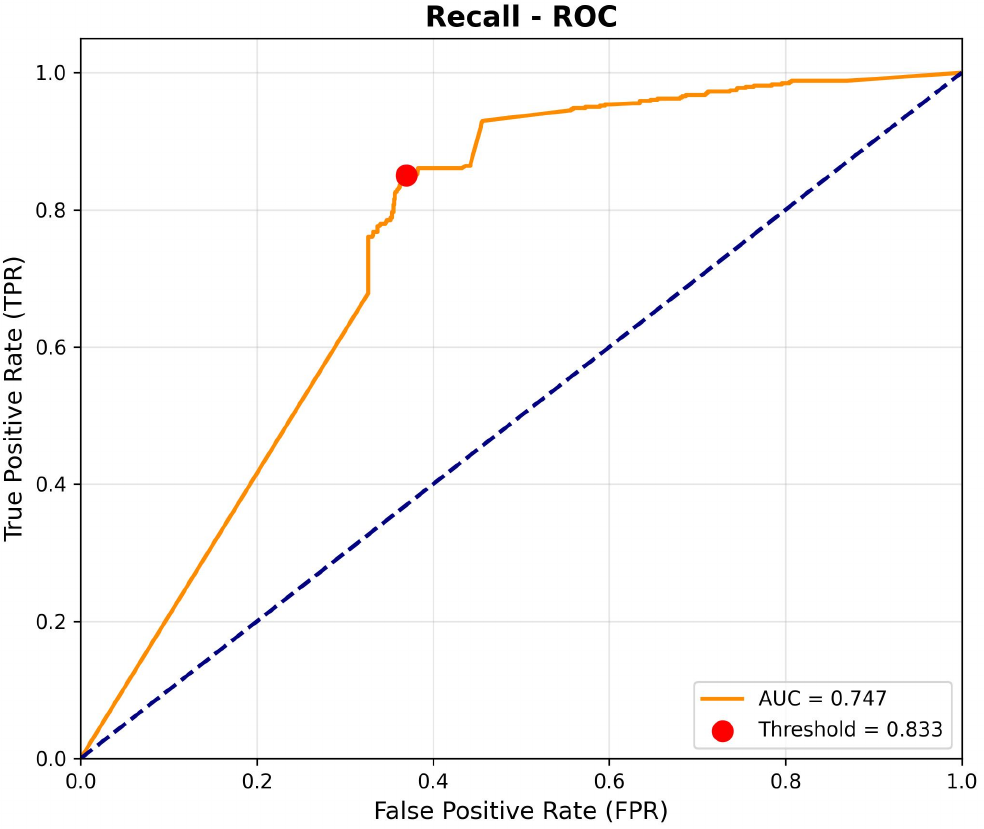}\\[-2pt]
    (d) Recall(ROC)
  \end{minipage}\hfill
  \begin{minipage}[b]{0.19\textwidth}
    \centering
    \includegraphics[width=\linewidth]{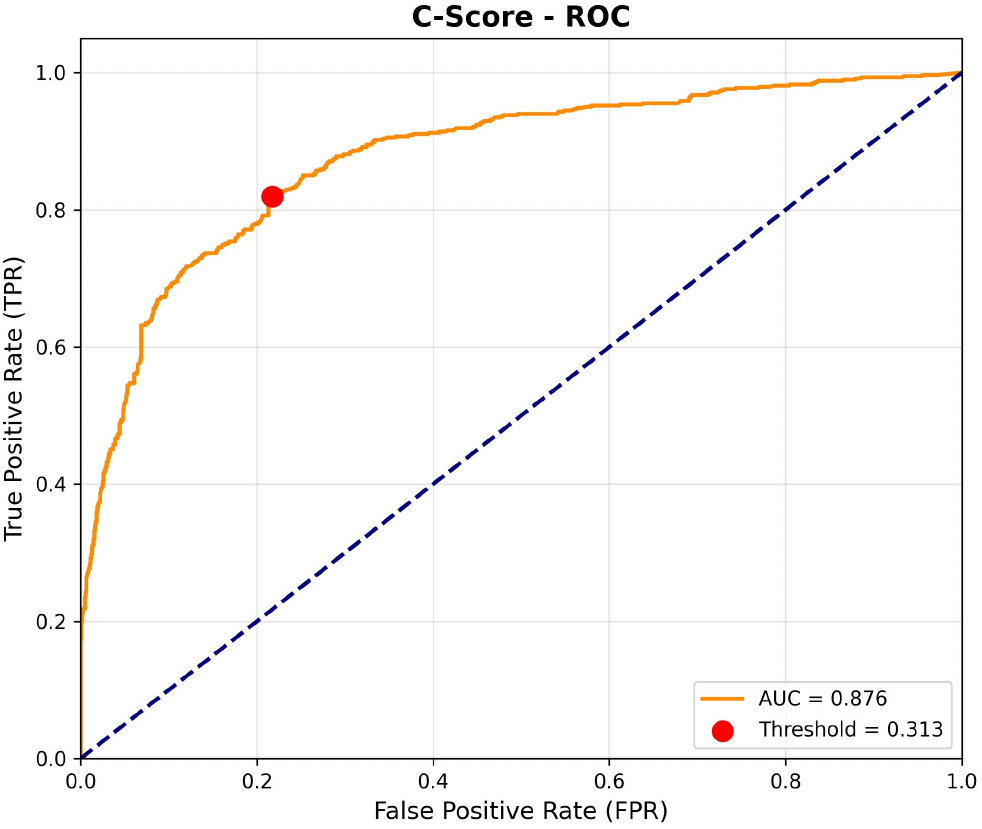}\\[-2pt]
    (e) C-Score(ROC)
  \end{minipage}

  \vspace{4pt}

  % -------- Row 2: Precision--Recall --------
  \begin{minipage}[b]{0.19\textwidth}
    \centering
    \includegraphics[width=\linewidth]{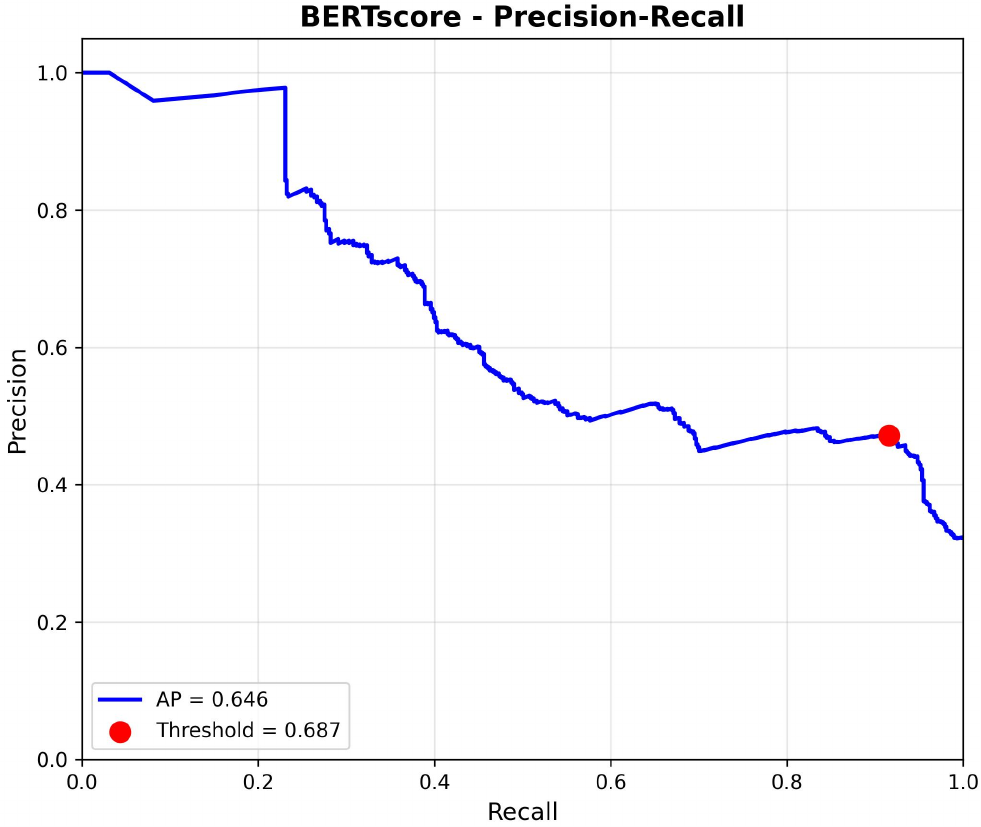}\\[-2pt]
    (f) BERTScore(PR)
  \end{minipage}\hfill
  \begin{minipage}[b]{0.19\textwidth}
    \centering
    \includegraphics[width=\linewidth]{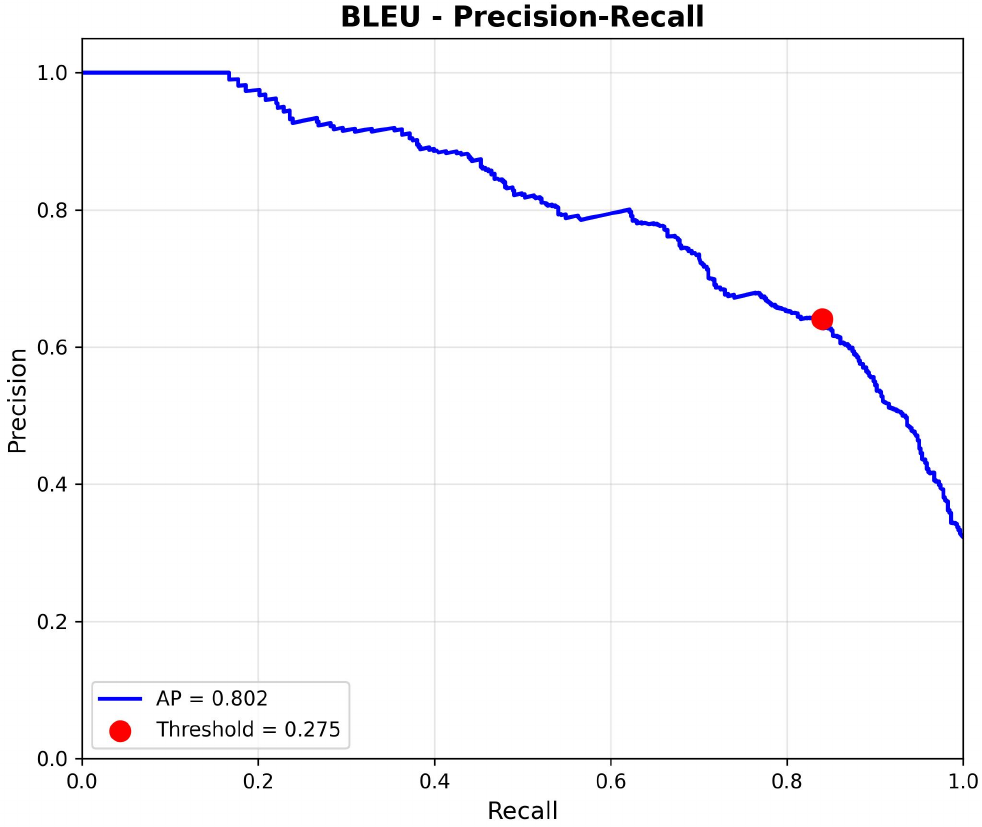}\\[-2pt]
    (g) BLEU(PR)
  \end{minipage}\hfill
  \begin{minipage}[b]{0.19\textwidth}
    \centering
    \includegraphics[width=\linewidth]{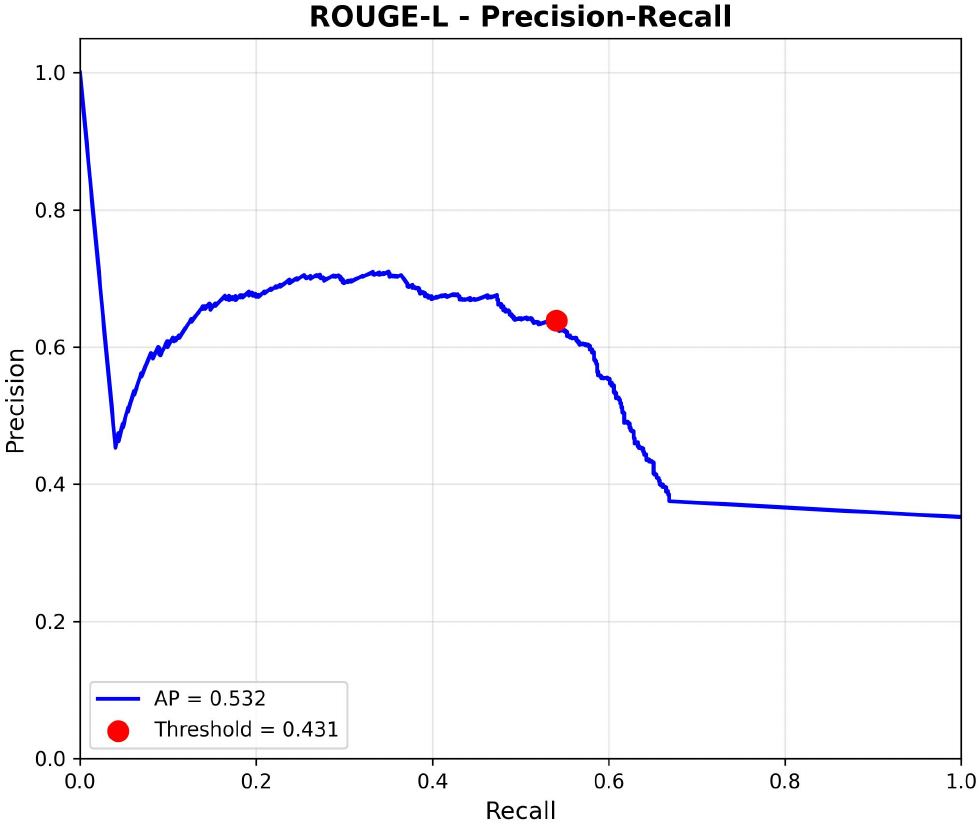}\\[-2pt]
    (h) ROUGE-L(PR)
  \end{minipage}\hfill
  \begin{minipage}[b]{0.19\textwidth}
    \centering
    \includegraphics[width=\linewidth]{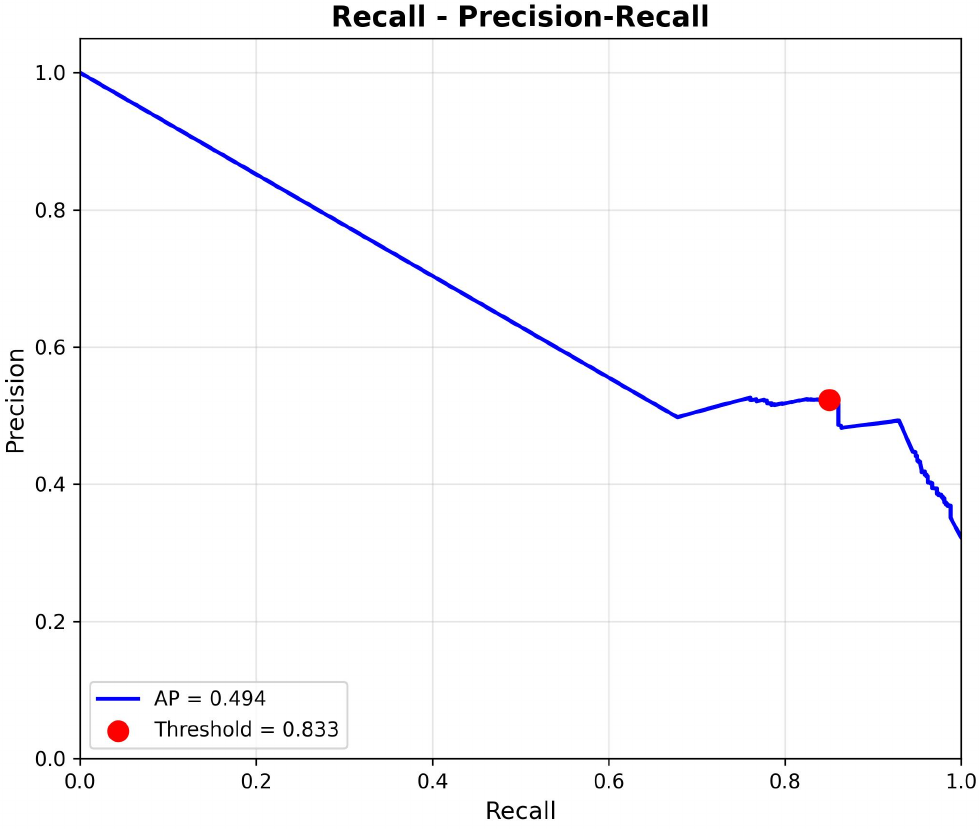}\\[-2pt]
    (i) Recall(PR)
  \end{minipage}\hfill
  \begin{minipage}[b]{0.19\textwidth}
    \centering
    \includegraphics[width=\linewidth]{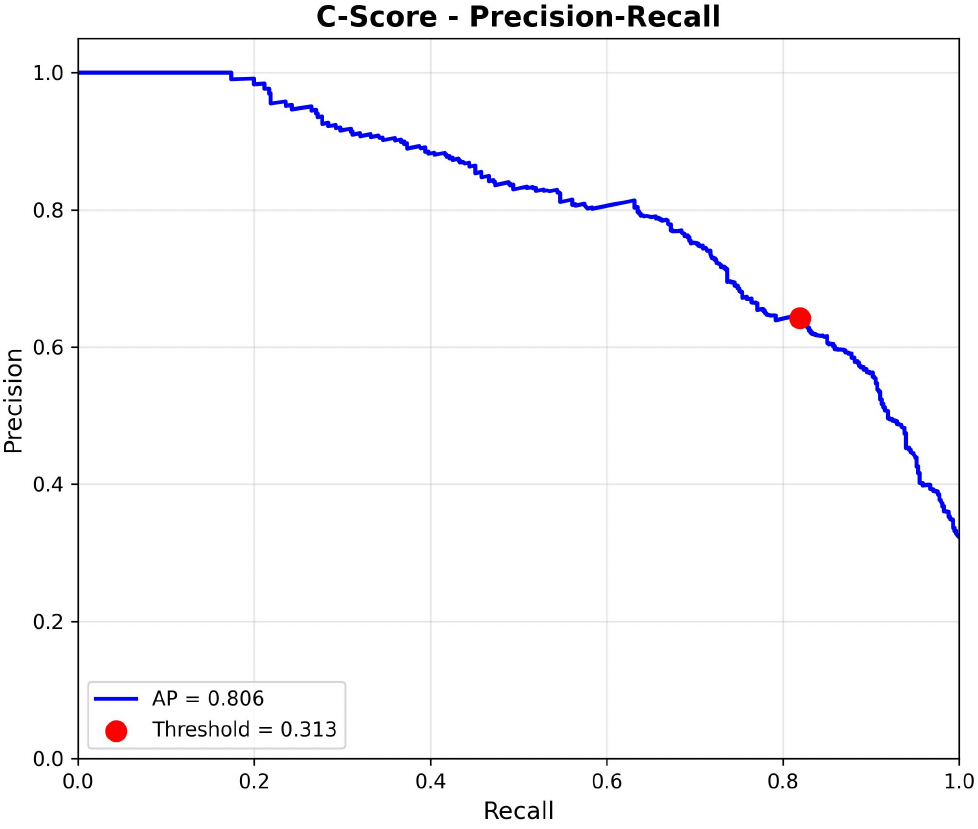}\\[-2pt]
    (j) C-Score(PR)
  \end{minipage}

  \caption{ROC curves (top row) and precision--recall curves (bottom row) for different metrics.}
  \label{fig:metric_curves}
\end{figure}

\end{document}